\documentclass[conference]{IEEEtran}

\usepackage{cite}
\usepackage{amsmath,amssymb,amsfonts}
\usepackage{algorithmic}
\usepackage{graphicx}
\usepackage{textcomp}
\usepackage{xcolor}

\usepackage{array}
\usepackage[hidelinks]{hyperref}
\usepackage{placeins}
\usepackage{dblfloatfix}

\ifCLASSOPTIONcompsoc
    \usepackage[caption=false,font=normalsize,labelfont=sf,textfont=sf]{subfig}
\else
    \usepackage[caption=false,font=footnotesize]{subfig}
\fi

\def\BibTeX{{\rm B\kern-.05em{\sc i\kern-.025em b}\kern-.08em
    T\kern-.1667em\lower.7ex\hbox{E}\kern-.125emX}}
\begin{document}

\title{Transformer-Based Vector Font Classification\\ Using Different Font Formats:\\ TrueType versus PostScript\\
}

\author{
    \IEEEauthorblockN{
        Takumu Fujioka\IEEEauthorrefmark{1} and Gouhei Tanaka\IEEEauthorrefmark{1}\IEEEauthorrefmark{2}
    }
    \IEEEauthorblockA{
        \IEEEauthorrefmark{1} Department of Computer Science, Nagoya Institute of Technology, Nagoya 466-8555, Japan
    }
    \IEEEauthorblockA{
        \IEEEauthorrefmark{2} International Research Center for Neurointelligence, The University of Tokyo, Tokyo 113-0033, Japan
    }
    \IEEEauthorblockA{
        t.fujioka.494@stn.nitech.ac.jp, gtanaka@nitech.ac.jp
    }
}

\maketitle

\begin{abstract}
    Modern fonts adopt vector-based formats, which ensure scalability without loss of quality.
    While many deep learning studies on fonts focus on bitmap formats, deep learning for vector fonts remains underexplored.
    In studies involving deep learning for vector fonts, the choice of font representation has often been made conventionally.
    However, the font representation format is one of the factors that can influence the computational performance of machine learning models in font-related tasks.
    Here we show that font representations based on PostScript outlines outperform those based on TrueType outlines in Transformer-based vector font classification.
    TrueType outlines represent character shapes as sequences of points and their associated flags, whereas PostScript outlines represent them as sequences of commands.
    In previous research, PostScript outlines have been predominantly used when fonts are treated as part of vector graphics, while TrueType outlines are mainly employed when focusing on fonts alone.
    Whether to use PostScript or TrueType outlines has been mainly determined by file format specifications and precedent settings in previous studies, rather than performance considerations.
    To date, few studies have compared which outline format provides better embedding representations.
    Our findings suggest that information aggregation is crucial in Transformer-based deep learning for vector graphics, as in tokenization in language models and patch division in bitmap-based image recognition models.
    This insight provides valuable guidance for selecting outline formats in future research on vector graphics.
\end{abstract}


\section{Introduction}

Just as images can be categorized into raster graphics and vector graphics, fonts are classified into bitmap fonts and vector fonts.
Vector fonts define the shapes of characters as geometric outlines, allowing them to scale without any loss in visual quality.
For this reason, most modern fonts adopt the vector format.
This suggests that, even in deep learning applications, fonts should ideally be handled in vector format rather than bitmap format.
However, the majority of deep learning research on fonts focus on bitmap representations.
To bridge this gap, research on deep learning specifically targeting vector fonts is necessary.

Vector fonts are not merely a subset of vector graphics.
Specifically, they do not contain information about stroke width or fill colors, and open paths are not allowed.
Additionally, the outer contours of vector fonts must follow a counterclockwise direction, while inner contours must follow a clockwise direction.
Furthermore, vector fonts have the distinctive property of being clearly separable into style and content.
Style refers to visual attributes such as stroke thickness, slant, and the presence or absence of serifs, while content refers to the character type and structure of the glyph.
Thus, vector fonts exhibit unique characteristics that are not present in general vector graphics, making research focused on vector fonts meaningful.

\begin{figure}[!t]
    \centerline{\includegraphics{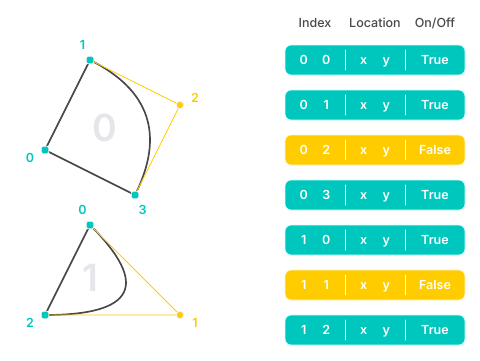}}
    \caption{
        TrueType outline representation.
        Each point is assigned an \textbf{index} to indicate which contour it belongs to and where it is located within that contour, and it is specified by its \textbf{location} \((x, y)\).
        The \textbf{on/off} flag determines whether a point is an on-curve point (\texttt{True}) or an off-curve control point (\texttt{False}).
        Off-curve points act as control points for quadratic Bézier curves, shaping the outline's curvature.
    }
    \label{fig:truetype_outline}
\end{figure}

\begin{figure}[!t]
    \centerline{\includegraphics{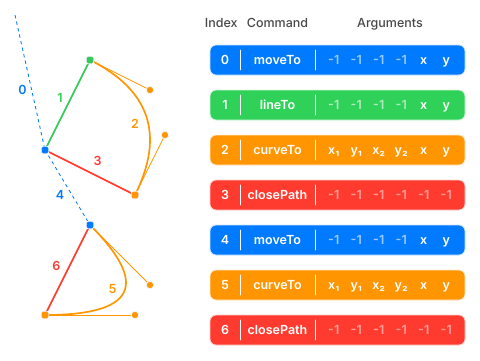}}
    \caption{
        PostScript outline representation.
        Each drawing step is identified by an \textbf{index}.
        The associated \textbf{command} specifies an operation such as \texttt{moveTo}, \texttt{lineTo}, \texttt{curveTo}, or \texttt{closePath}, with parameters defining coordinates, including control points for Bézier curves and end-points.
    }
    \label{fig:postscript_outline}
\end{figure}

Vector font outlines are represented in two primary formats: TrueType outlines and PostScript outlines.
TrueType outlines, as shown in Figure~\ref{fig:truetype_outline}, are used in TrueType fonts \cite{apple_truetype} and OpenType fonts \cite{microsoft_opentype} that adopt the TrueType format.
They represent shapes as a sequence of points and flags, utilizing quadratic Bézier curves to define curves.
On the other hand, PostScript outlines, illustrated in Figure~\ref{fig:postscript_outline}, are employed in PostScript fonts \cite{adobe_type1} and OpenType fonts \cite{microsoft_opentype} that adopt the PostScript format.
These outlines use a sequence of commands based on the PostScript language, with cubic Bézier curves for defining curves.
Consequently, shapes expressed in the TrueType outline format are a subset of those representable in the PostScript outline format.
Moreover, Scalable Vector Graphics (SVG), a major file format for vector graphics, adopts a format similar to PostScript outlines, as it also uses cubic Bézier curves for curve representation.

Vector graphics represent a form of data that lies between texts and images.
Existing deep learning research on vector graphics can be broadly categorized into two approaches: one inspired by image generation techniques and the other influenced by language models.
Among these, DeepSVG \cite{carlier2020deepsvg} stands out as a representative model that utilizes Transformer-based architectures \cite{vaswani2017attention} for vector graphics generation.
Due to its focus on SVG-format data, DeepSVG employs sequence-based embedding representations derived from drawing commands.
Several subsequent studies on vector fonts \cite{wang2020attribute2font, aoki2022svg}, building upon DeepSVG, have adopted the same sequence-based embedding approach.
On the other hand, works like TrueType Transformer ($T^3$) \cite{nagata2022truetype, nagata2023contour} are inspired by BERT \cite{devlin2019bert}, a language model.
These works adapt BERT's mechanisms to vector fonts for tasks such as font classification and path data completion.
Notably, these approaches rely on embeddings based on TrueType outlines.
One reason for this preference may be that the largest dataset provider, Google Fonts \cite{googlefonts}, primarily offers TrueType fonts \cite{apple_truetype}.
As a result, two distinct outline formats are used separately in deep learning for vector fonts.
It appears that the choice of outline format is influenced more by historical research trends and file format specifications than by performance considerations.
To date, few studies have systematically compared the effectiveness of embedding representations between these two outline formats.

In this study, we compare TrueType outlines and PostScript outlines in the performance of font classification tasks using Transformer-based models.
Our architecture adopts a classification token (CLS) approach, similar to those used in BERT \cite{devlin2019bert}, Vision Transformer (ViT), \cite{dosovitskiy2021an} and $T^3$ \cite{nagata2022truetype}.
We conducted experiments on tasks such as classifying fonts with complex shapes like Kanji characters and categorizing font weights.
Our contributions are summarized as follows:

\begin{itemize}
    \item We show that Transformer-based vector font classification models can be effectively applied to fonts containing complex shapes, such as Kanji characters.
    \item We demonstrate that embedding representations based on PostScript outlines outperform those based on TrueType outlines in deep learning tasks involving vector fonts with Transformers.
    \item We reveal that this performance difference is primarily due to the segmentation process, where transforming point sequences into command sequences improves representation quality.
\end{itemize}

\section{Related Work}

\begin{table}[!b]
    \renewcommand{\arraystretch}{1.3}
    \caption{
        \normalfont
        Outline formats used in previous vector font studies.
        These representations are categorized into command sequences and point sequences.
        In the context of vector fonts, these correspond to PostScript outlines and TrueType outlines.
    }
    \label{tab:study_outline_format}
    \centering
    \begin{tabular}{ll}
        \hline
        \textbf{Study}                              & \textbf{Outline formats} \\
        \hline
        SVG-VAE \cite{lopes2019svg}                 & Subset of SVG Commands   \\
        Im2Vec \cite{reddy2021im2vec}               & Points                   \\
        DeepSVG \cite{carlier2020deepsvg}           & Subset of SVG Commands   \\
        Aoki and Aizawa \cite{aoki2022svg}          & Subset of SVG Commands   \\
        IconShop \cite{wu2023iconshop}              & Subset of SVG Commands   \\
        DeepVecFont \cite{wang2021deepvecfont}      & Custom Drawing Commands  \\
        DeepVecFont-v2 \cite{wang2023deepvecfontv2} & Custom Drawing Commands  \\
        $T^3$ \cite{nagata2022truetype}             & TrueType Points          \\
        Nagata et al. \cite{nagata2023contour}      & TrueType Points          \\
        \hline
    \end{tabular}
\end{table}

\begin{table*}[!b]
    \renewcommand{\arraystretch}{1.3}
    \caption{
        \normalfont
        Drawing commands used in vector font outlines.
        The commands \texttt{moveTo}, \texttt{lineTo}, and \texttt{closePath} are common to both PostScript and TrueType outlines.
        The command \texttt{qCurveTo} is used only in TrueType outlines, whereas \texttt{curveTo} is exclusive to PostScript outlines.
    }
    \label{tab:drawing_commands}
    \centering
    \begin{tabular}{>{\centering\arraybackslash}m{2cm}>{\centering\arraybackslash}m{3cm}m{6cm}>{\centering\arraybackslash}m{4cm}}
        \hline
        \textbf{Command}   & \textbf{Arguments}           & \textbf{Description}                                                                                    & \textbf{Visualization}                                 \\
        \hline
        \texttt{moveTo}    & \(x, y\)                     & Move the cursor to the end-point \((x, y)\) without drawing anything.                                   & \includegraphics[scale=1]{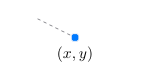}    \\
        \hline
        \texttt{lineTo}    & \(x, y\)                     & Draw a line to the point \((x, y)\).                                                                    & \includegraphics[scale=1]{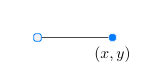}    \\
        \hline
        \texttt{qCurveTo}  & \(x_1, y_1, x, y\)           & Draw a quadratic Bézier curve with control point \((x_1, y_1)\) and end-point \((x, y)\).               & \includegraphics[scale=1]{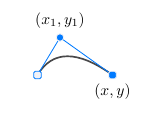} \\
        \hline
        \texttt{curveTo}   & \(x_1, y_1, x_2, y_2, x, y\) & Draw a cubic Bézier curve with control points \((x_1, y_1)\), \((x_2, y_2)\), and end-point \((x, y)\). & \includegraphics[scale=1]{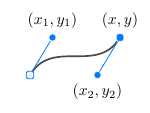}   \\
        \hline
        \texttt{closePath} & \(\varnothing\)              & Close the path by moving the cursor back to the path’s starting position.                               & \includegraphics[scale=1]{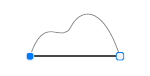} \\
        \hline
    \end{tabular}
\end{table*}

\subsection{Transformer}

Transformer \cite{vaswani2017attention} is a neural network architecture that has achieved remarkable success in the field of natural language processing (NLP).
By utilizing the attention mechanism, Transformer enables the processing of variable-length sequential data.

Originally proposed as an Encoder-Decoder model for translation tasks, Transformer has since evolved into models that use only the Encoder or the Decoder.
BERT \cite{devlin2019bert} is a language model that utilizes only the Encoder of the Transformer.
By combining pre-training and fine-tuning, BERT has achieved state-of-the-art performance in tasks such as question answering and document classification.
BERT encodes the entire input sequence into a fixed-length vector using the classification token (CLS), which is added to the beginning of the input sequence.
The encoded CLS token is then used for classification tasks.

Transformer has also been successfully applied in the field of computer vision.
Vision Transformer (ViT) \cite{dosovitskiy2021an} processes images by dividing them into multiple patches and treating each patch as a token, enabling effective image analysis.

Since vector graphics are represented as variable-length sequential data, Transformer provides a suitable framework for deep learning of vector graphics.
Furthermore, vector graphics can be considered an intermediate data form between texts and images, as they are images constructed from variable-length sequences.
Therefore, vector graphics can leverage insights from both NLP and computer vision applications of Transformer.

\subsection{Deep Learning for Vector Font}

Vector graphics are represented as variable-length sequential data, making it challenging to handle them in deep learning research until recently.
SVG-VAE \cite{lopes2019svg} is one of the earliest models to address vector graphics generation.
It generates vector graphics from bitmap images using Variational Autoencoder (VAE) \cite{kingma2022auto}, leveraging an embedding representation based on sequences of SVG commands.
However, its applications were limited to Latin font generation.

Im2Vec \cite{reddy2021im2vec} also generates vector graphics from bitmap images but adopts an embedding representation similar to TrueType outlines, using sequences of points and their corresponding flags.
Unlike SVG-VAE, Im2Vec extends its applicability not only to Latin fonts but also to emojis and icons.

DeepSVG \cite{carlier2020deepsvg} is the first model that apply Transformers to vector graphics generation.
By combining a VAE \cite{kingma2022auto} with a hierarchical Transformer architecture and feed-forward prediction, DeepSVG can successfully generate vector graphics.
It utilizes an embedding representation based on command sequences and conducts experiments on icons and Latin fonts.
DeepSVG has inspired several subsequent studies.
Aoki and Aizawa \cite{aoki2022svg} extended DeepSVG by focusing on font generation and introducing AdaIN \cite{huang2017arbitrary} and Chamfer Loss, which enabled successful generation of complex shapes like Kanji characters.
IconShop \cite{wu2023iconshop}, another extended work, specializes in icon generation.
These studies maintain the same command-sequence-based embedding representation as DeepSVG.

DeepVecFont \cite{wang2021deepvecfont} is a model that utilizes both bitmap and vector modalities for font generation.
For generating vector graphics, it employs an LSTM-based approach \cite{hochreiter1997lstm}.
DeepVecFont adopts an embedding representation based on sequences of drawing commands, with coordinates expressed in relative values.
DeepVecFont-v2 \cite{wang2023deepvecfontv2}, an improved version of DeepVecFont, introduces several enhancements,
including replacing LSTM with Transformer, representing coordinates in absolute values, and incorporating additional information about the starting point of commands.

$T^3$ \cite{nagata2022truetype} is a model focused on vector font classification.
It successfully adapts the BERT model architecture to vector fonts, enabling tasks such as character recognition and font style classification directly from vector format data.
Nagata et al. \cite{nagata2023contour} conducted research on contour completion for vector graphics.
Both studies adopt embedding representations based on TrueType outlines.

The studies discussed above adopt different outline representations.
Table~\ref{tab:study_outline_format} summarizes the outline formats used in previous vector font studies.
These outline representations can be broadly categorized into command sequences and point sequences.
In the context of vector fonts, these correspond to PostScript outlines, which use drawing commands, and TrueType outlines, which represent contours as a set of points with flags.

\begin{figure*}[!t]
    \centerline{\includegraphics{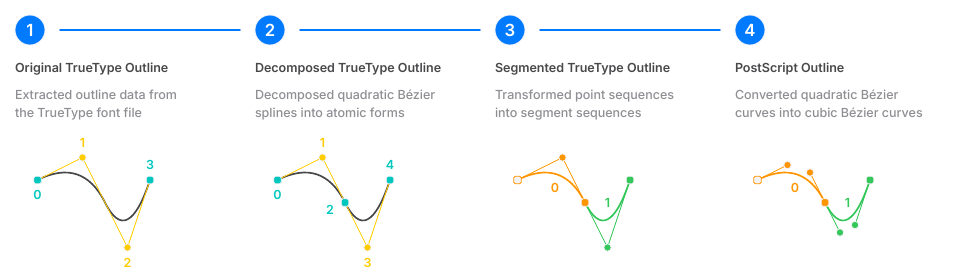}}
    \caption{
        Process of converting a TrueType outline into a PostScript outline.
        This transformation involves multiple steps, including decomposing quadratic Bézier splines, restructuring point sequences into segments, and converting quadratic Bézier curves into cubic Bézier curves.
        Each stage is illustrated in the figure, showing how the outline evolves through the transformation.
    }
    \label{fig:tt_to_ps}
\end{figure*}

\section{Method}

\subsection{Outline Formats}

Vector font outlines are mainly categorized into two types: PostScript outlines and TrueType outlines.
A PostScript outline consists of four types of drawing commands: \texttt{moveTo}, \texttt{lineTo}, \texttt{curveTo}, and \texttt{closePath}.
A TrueType outline is originally represented as a sequence of points and their flags, and it can be converted into drawing commands.
While a PostScript outline represents curves using cubic Bézier curves, a TrueType outline uses quadratic Bézier curves. Therefore, in the command representation of a TrueType outline, \texttt{qCurveTo} is used instead of \texttt{curveTo}.
Table~\ref{tab:drawing_commands} summarizes the drawing commands used in PostScript and TrueType outlines.

In this study, we compare TrueType outlines with their corresponding PostScript outlines obtained through conversion.
Since TrueType outlines are a subset of PostScript outlines, converting a PostScript outline into a TrueType outline requires approximation, leading to information loss and making a fair comparison difficult.
Figure~\ref{fig:tt_to_ps} illustrates the process of converting a TrueType outline into a PostScript outline.
In our experiments, we compare performance across the four patterns shown in Figure~\ref{fig:tt_to_ps}.
TrueType outlines may omit intermediate on-curve points, known as implicit on-curve points, in consecutive curves, requiring these points to be reconstructed when converting into a command sequence.
This reconstruction process may affect performance.
Additionally, when converting a command sequence based on TrueType outlines into one based on PostScript outlines, the number of control points increases by one, which may also impact performance.
To evaluate the impact of each transformation process on performance, we conduct experiments that include these transformation processes and compare performance across four different patterns.

\subsection{Outline Embedding}

The original TrueType outline and decomposed TrueType outline are embedded following the method of $T^3$ \cite{nagata2022truetype}.
As shown in Figure~\ref{fig:truetype_outline}, a TrueType outline is represented as a sequence of points and their corresponding flags.
Each point is characterized by a contour index, a point index, its coordinates \((x, y)\), and an on/off flag.
A point \( P^{i}_{j} \) is represented as a five-dimensional vector as follows:
\begin{equation}
    P^{i}_{j} = (i, j, x^{i}_{j}, y^{i}_{j}, o^{i}_{j}),
    \label{eq:point_representation_tt}
\end{equation}
where \( i \in \{1, \dots, N_c\} \) is the contour index, with \( N_c \) denoting the number of contours in the character outline,
\( j \in \{1, \dots, N_{p_i}\} \) represents the point index indicating the order within contour \( i \), with \( N_{p_i} \) denoting the number of points in contour \( i \),
\( (x^{i}_{j}, y^{i}_{j}) \in \mathbb{R}^2 \) specifies the point's coordinates,
and \( o^{i}_{j} \in \{0, 1\} \) is the on/off flag, with \( o^{i}_{j} = 1 \) for on-curve points and \( o^{i}_{j} = 0 \) for off-curve control points.

The embedding vector for each point is obtained by summing the embeddings of its individual components:
\begin{equation}
    e^{i}_{j} = e_{\text{c\_idx}}(i) + e_{\text{p\_idx}}(j) + e_{\text{loc}}(x^{i}_{j}, y^{i}_{j}) + e_{\text{flag}}(o^{i}_{j}),
    \label{eq:embedding_representation_tt}
\end{equation}
where \( e_{\text{c\_idx}}(i) \) is the embedding of the contour index,
\( e_{\text{p\_idx}}(j) \) is the embedding of the point index within the contour,
\( e_{\text{loc}}(x^{i}_{j}, y^{i}_{j}) \) is the embedding of the point's coordinates,
and \( e_{\text{flag}}(o^{i}_{j}) \) is the embedding of the on/off flag.

The segmented TrueType outline and PostScript outline are embedded following the method of DeepSVG \cite{carlier2020deepsvg}.
As shown in Figure~\ref{fig:postscript_outline}, an outline is represented as a sequence of drawing commands.
Each command \( C_i \) is represented as follows:
\begin{equation}
    C_i = (i, c_i, X_i),
    \label{eq:command_representation}
\end{equation}
where \( i \) represents the command index, \( c_i \) represents the command type, and \( X_i \) is the set of coordinate arguments.
In the case of the PostScript outline, the command type and coordinate arguments are given by
\begin{equation}
    c_i \in \{\texttt{moveTo}, \texttt{lineTo}, \texttt{curveTo}, \texttt{closePath}\},
\end{equation}
\begin{equation}
    X_i = (x^i_1, y^i_1, x^i_2, y^i_2, x^i, y^i) \in \mathbb{R}^6.
\end{equation}
For the segmented TrueType outline, which uses quadratic Bézier curves instead of cubic ones, the command type and coordinate arguments are represented as follows:
\begin{equation}
    c_i \in \{\texttt{moveTo}, \texttt{lineTo}, \texttt{qCurveTo}, \texttt{closePath}\},
\end{equation}
\begin{equation}
    X_i = (x^i_1, y^i_1, x^i, y^i) \in \mathbb{R}^4.
\end{equation}
Since quadratic Bézier curves require only one control point, the segmented TrueType outline omits \((x^i_2, y^i_2)\).
Unused arguments are set to \(-1\) for padding.

The embedding vector for each command is obtained by summing the embeddings of its components as follows:
\begin{equation}
    e_i = e_{\text{idx}}(i) + e_{\text{cmd}}(c_i) + e_{\text{args}}(X_i),
    \label{eq:command_embedding}
\end{equation}
where \( e_{\text{idx}}(i) \) is the embedding of the command index,
\( e_{\text{cmd}}(c_i) \) is the embedding of the command type,
and \( e_{\text{args}}(X_i) \) is the embedding of the coordinate arguments.

\subsection{Model Architecture}

\begin{figure}[!t]
    \centerline{\includegraphics{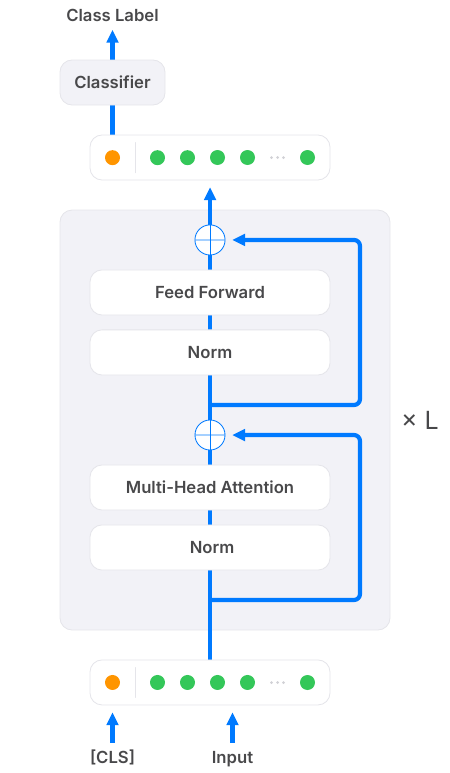}}
    \caption{
        Model architecture for font classification.
        The input sequence, consisting of vector font outline data, is first mapped to embedding vectors.
        A classification token (CLS) is prepended to the sequence before being processed by the Transformer Encoder.
        The encoder consists of multiple layers of multi-head self-attention, feed-forward networks, and layer normalization.
        The output corresponding to the CLS token is passed to a classifier to predict the font category.
    }
    \label{fig:model_architecture}
\end{figure}

Figure~\ref{fig:model_architecture} illustrates the model architecture used in this study.
The model adopts a Transformer Encoder-based architecture, similar to BERT \cite{devlin2019bert}, ViT \cite{dosovitskiy2021an}, and $T^3$ \cite{nagata2022truetype}.

The input consists of vector font outline data, which are represented as a sequence of points or commands.
Let \( N \) be the length of the input sequence and \( D \) be the embedding dimension.
Each element in the sequence is mapped to a \( D \)-dimensional embedding vector through an embedding layer.
A classification token (CLS) is prepended to the sequence, resulting in an expanded sequence of length \( N + 1 \), which is then fed into the Transformer Encoder.

The Transformer Encoder consists of \( L \) layers, where each layer applies a self-attention mechanism followed by a feed-forward network.
Each self-attention layer employs multi-head attention with \( H \) attention heads.
The final output of the Transformer Encoder is extracted from the position corresponding to the CLS token.
This vector serves as a summary representation of the entire input.
It is then passed through a fully connected layer to produce the final class prediction among \( K \) possible classes.
Cross-entropy loss is used as a loss function.

\section{Experiments}

\subsection{Dataset}

The dataset consists of font data collected from Google Fonts \cite{googlefonts}.
For the font style classification experiments, we selected 16 Japanese fonts from Google Fonts, prioritizing those with high relevance.
Monospaced fonts were excluded, as their fixed-width design makes them unsuitable for this task.
All fonts were set to the Regular weight.

For the font weight classification experiments, we selected four Japanese fonts from Google Fonts, again prioritizing those with high relevance.
We used four weight variations: Light, Regular, Medium, and Bold, forming a dataset with 16 classes in total.

All available characters from each font were included, resulting in approximately 3,000 to 8,000 characters per font.
The dataset was randomly split into training, validation, and test sets for each font, with proportions of 80\%, 10\%, and 10\%, respectively.

\begin{figure*}[!t]
    \centering
    \subfloat[Original TrueType Outline]{
        \includegraphics[width=0.48\textwidth]{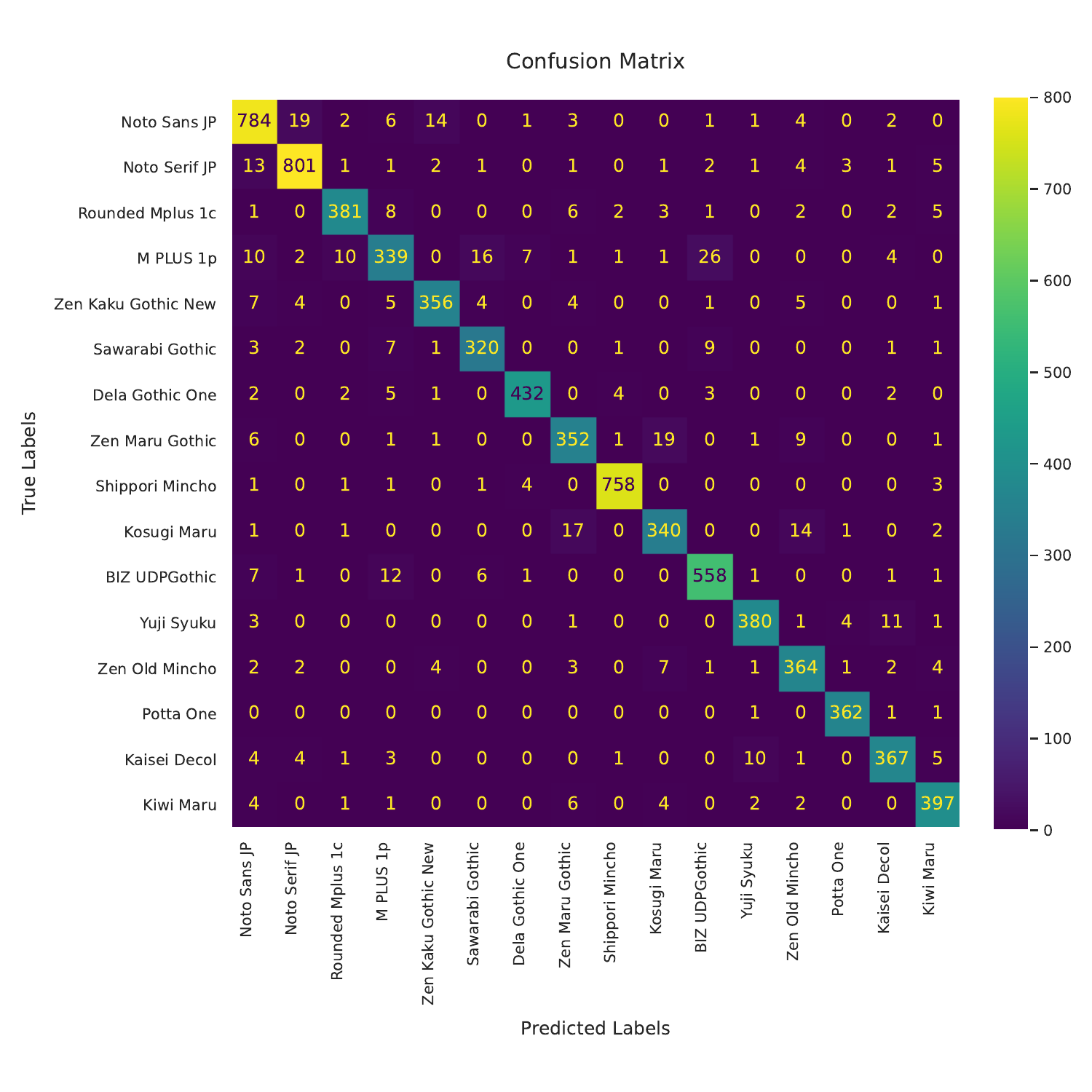}
    }
    \hfill
    \subfloat[PostScript Outline]{
        \includegraphics[width=0.48\textwidth]{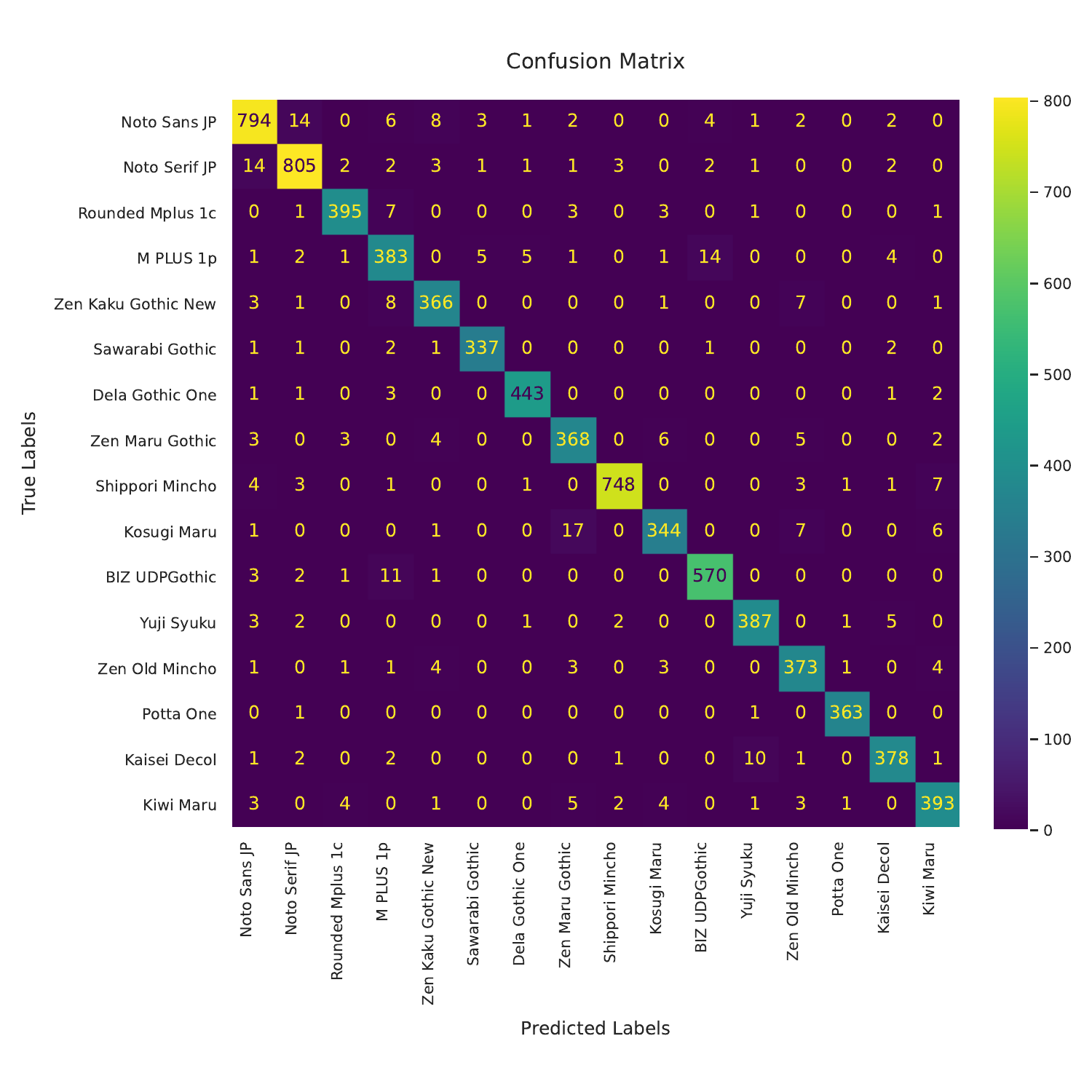}
    }
    \caption{
        Confusion matrices for font style classifications with (a) original TrueType outline and (b) PostScript outline.
        It can be observed that the PostScript outline performs slightly better than the original TrueType outline.
    }
    \label{fig:style_confusion_matrices}
\end{figure*}

\subsection{Implementation Details}

The same hyperparameter settings were used for both font style classification and font weight classification.
The embedding dimension \( D \) was set to 64, the hidden dimension of the feed-forward network to 128, the number of attention heads \( H \) to 4, and the number of encoder layers \( L \) to 3.
The model was trained for 512 epochs using the AdamW \cite{loshchilov2018decoupled} optimizer with a learning rate of \( 1 \times 10^{-4} \) and a batch size of 1024.
The learning rate was scheduled with a warm-up phase for the first 250 steps, followed by decay according to an inverse square root schedule.
Training was performed on two NVIDIA RTX 6000 Ada GPUs.

\subsection{Font Style Classification}

Figure~\ref{fig:style_confusion_matrices} shows the confusion matrices for font style classification on the test set for original TrueType outline and PostScript outline.
Table~\ref{tab:style_performance} summarizes the scores of four evaluation metrics for font style classification on the test set.
The results indicate that PostScript outlines outperform TrueType outlines across all evaluation metrics.

\begin{table}[!t]
    \renewcommand{\arraystretch}{1.3}
    \caption{
        \normalfont
        Performance comparison of different outline formats for font style classification.
        The results indicate that the PostScript outline outperforms the TrueType outlines, primarily due to segmentation.
    }
    \label{tab:style_performance}
    \centering
    \begin{tabular}{lcccc}
        \hline
        \textbf{Outline}    & \textbf{Loss} & \textbf{Acc.} & \textbf{Macro F1} & \textbf{W-F1} \\
        \hline
        Original TrueType   & 0.2027        & 93.7\%        & 93.3\%            & 93.7\%        \\
        Decomposed TrueType & 0.2033        & 93.3\%        & 92.8\%            & 93.3\%        \\
        Segmented TrueType  & 0.1313        & 95.3\%        & 95.1\%            & 95.3\%        \\
        PostScript          & 0.1156        & 95.7\%        & 95.6\%            & 95.7\%        \\
        \hline
    \end{tabular}
\end{table}

Comparing the four outline formats, the performance gap between decomposed TrueType outline and segmented TrueType outline is most significant.
Thus, segmentation appears to be a major contributing factor to the performance difference between TrueType outlines and PostScript outlines.

\begin{figure}[!t]
    \centering
    \subfloat[Training Loss]{
        \includegraphics[width=0.45\columnwidth]{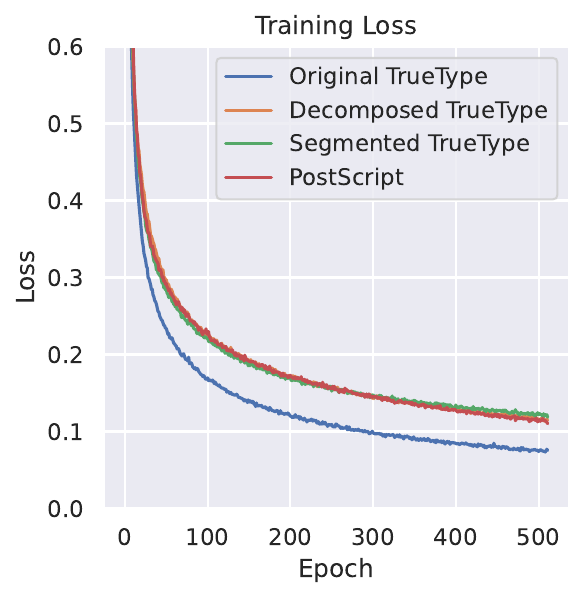}
    }
    \hfill
    \subfloat[Validation Loss]{
        \includegraphics[width=0.45\columnwidth]{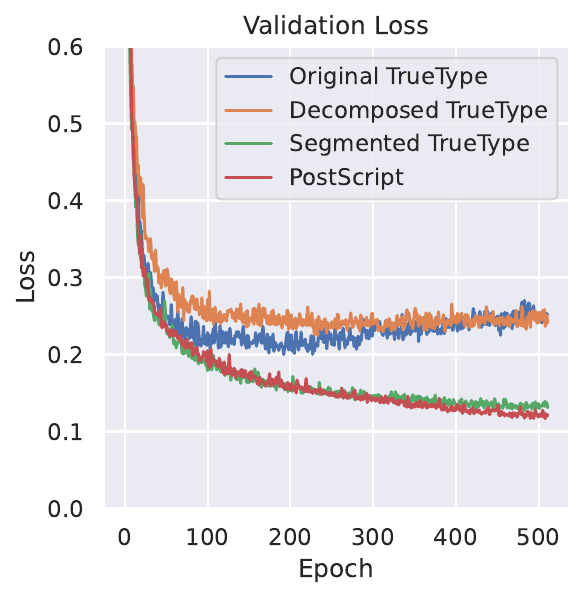}
    }
    \caption{
        Training and validation losses for font style classification across different outline formats.
        The original TrueType outline and decomposed TrueType outline tend to overfit the training data, whereas the segmented TrueType outline and PostScript outline exhibit better generalization performance.
    }
    \label{fig:style_loss}
\end{figure}

Figure~\ref{fig:style_loss} presents the training and validation losses for all four outline formats.
Overfitting is observed for original TrueType outline and decomposed TrueType outline, but not for segmented TrueType outline and PostScript outline.

\subsection{Font Weight Classification}

\begin{figure*}[!t]
    \centering
    \subfloat[Original TrueType Outline]{
        \includegraphics[width=0.48\textwidth]{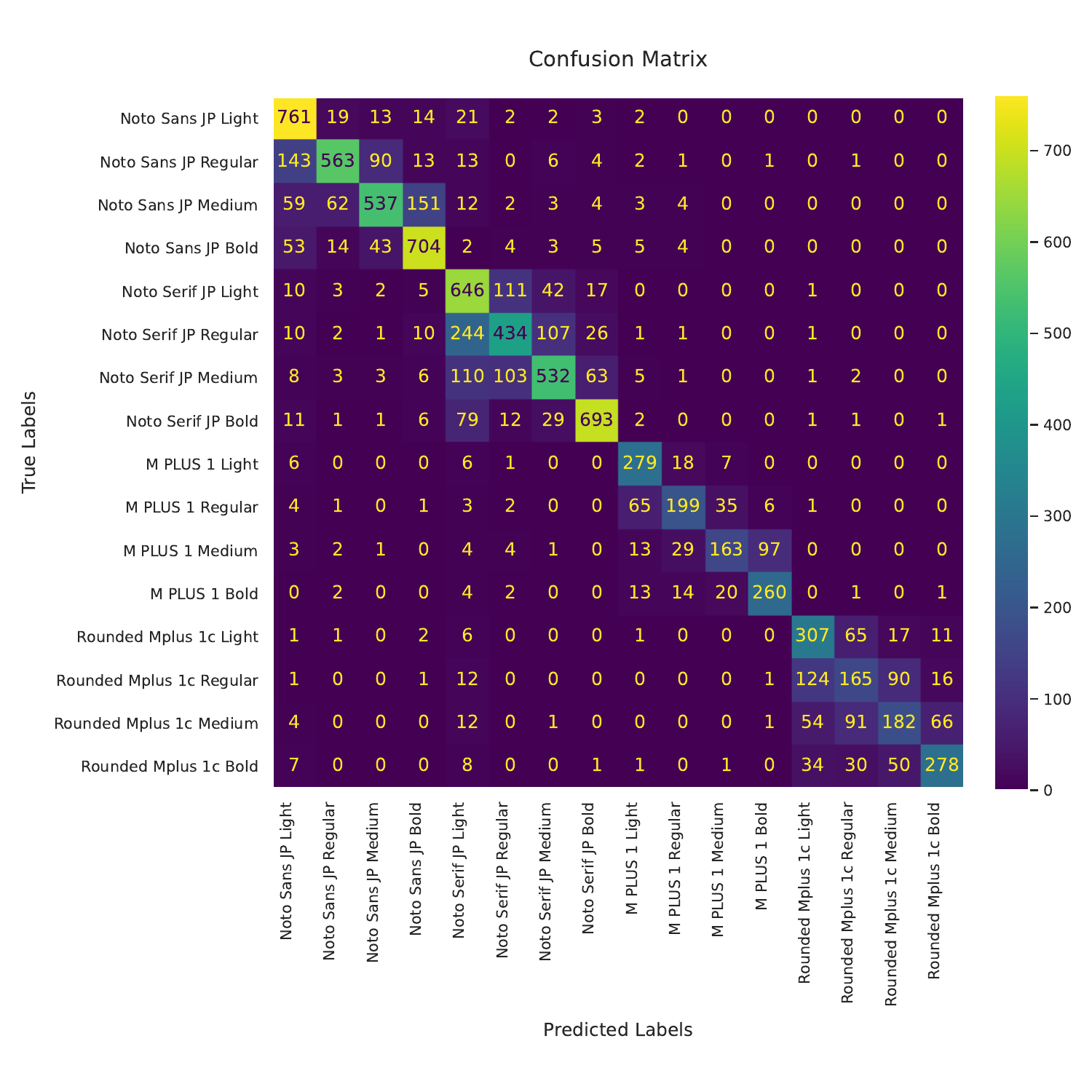}
    }
    \hfill
    \subfloat[PostScript Outline]{
        \includegraphics[width=0.48\textwidth]{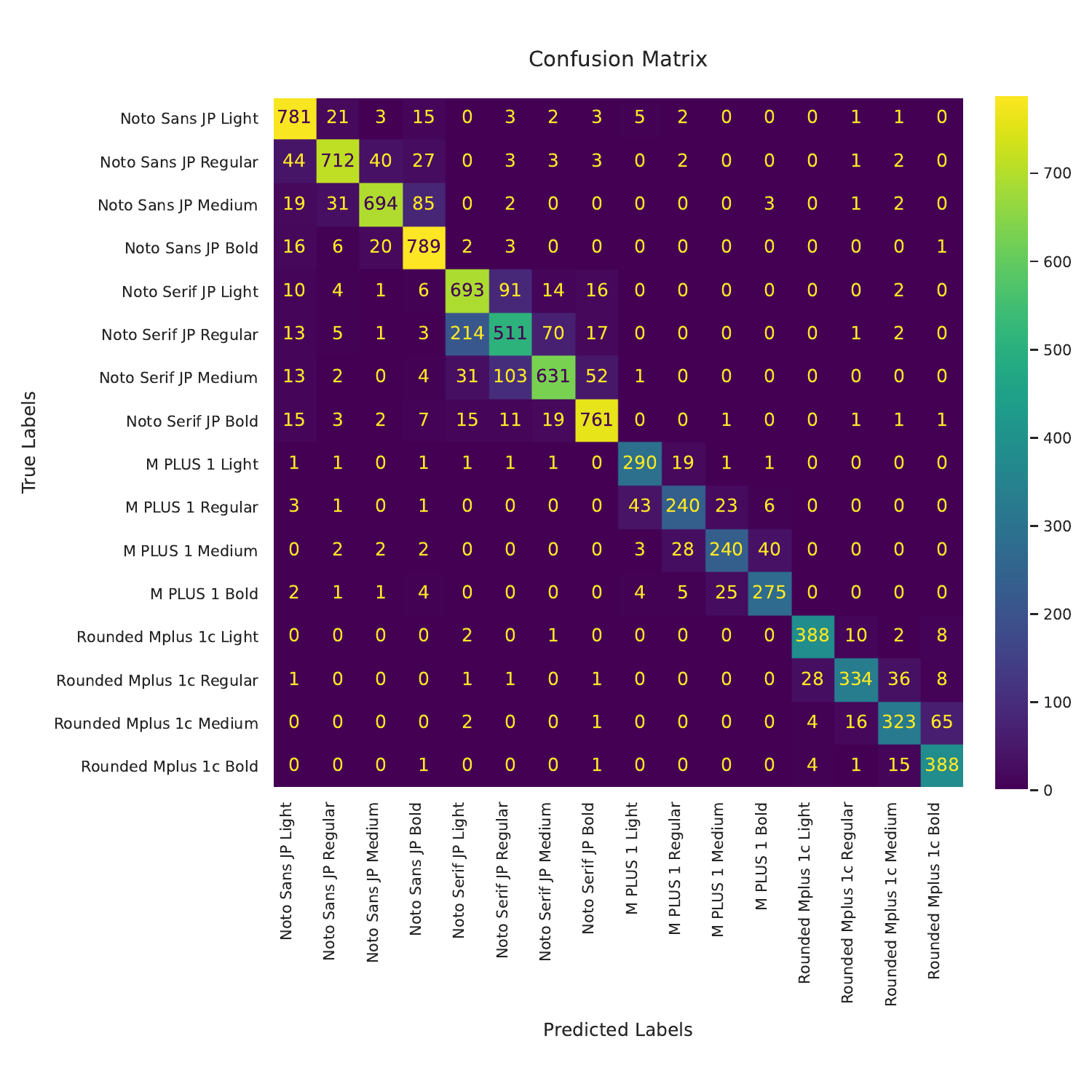}
    }
    \caption{
        Confusion matrices for font weight classification with (a) original TrueType outline and (b) PostScript outline.
        The results suggest that the PostScript outline outperforms the TrueType outline.
    }
    \label{fig:weight_confusion_matrix}
\end{figure*}

Figure~\ref{fig:weight_confusion_matrix} shows the confusion matrices for font weight classification on the test set for original TrueType outline and PostScript outline.
Table~\ref{tab:weight_performance} summarizes the scores of four evaluation metrics for font weight classification on the test set.
The results of font weight classification also demonstrate that PostScript outlines outperform TrueType outlines.

In the case of font weight classification, the performance gap between decomposed TrueType outline and segmented TrueType outline is remarkable.
Thus, the impact of segmentation is suggested to be significant in font weight classification.

Decomposed TrueType outline exhibits lower performance than original TrueType outline.
The increased sequence length, resulting from the reconstruction of implicit on-curve points, likely makes it more challenging to capture global information.

PostScript outline outperforms segmented TrueType outline.
The use of two control points in PostScript outlines for curve representation facilitates the learning of complex shapes, contributing to the improved performance.

Overall, font weight classification demonstrates lower performance compared to font style classification.
This trend suggests that font weight classification is inherently more challenging.
The confusion matrices in Figure~\ref{fig:weight_confusion_matrix} further illustrate this difficulty, showing a high degree of misclassification among similar font weights.

\begin{figure}[!t]
    \centering
    \subfloat[Training Loss]{
        \includegraphics[width=0.45\columnwidth]{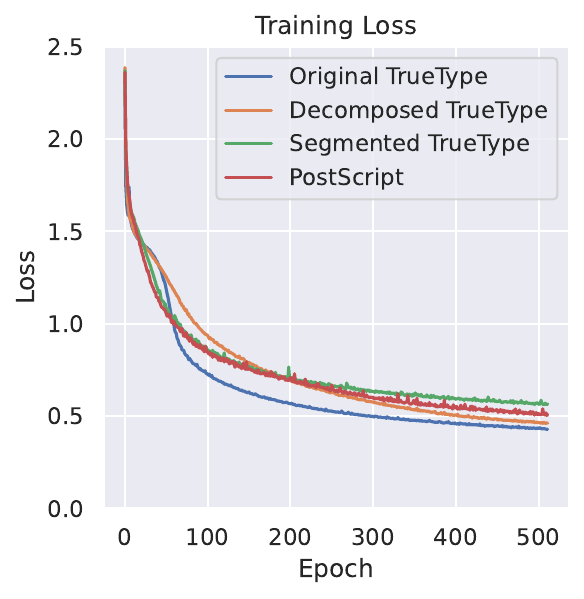}
    }
    \hfill
    \subfloat[Validation Loss]{
        \includegraphics[width=0.45\columnwidth]{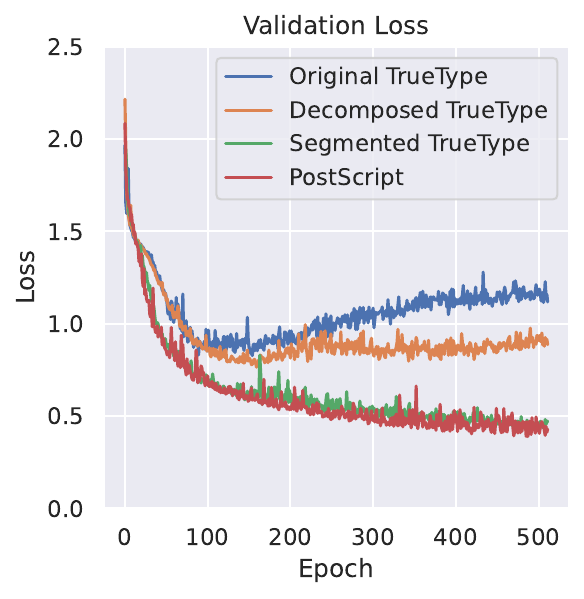}
    }
    \caption{
        Training and validation losses for font weight classification across different outline formats.
        The original TrueType outline and decomposed TrueType outline show signs of overfitting, while the segmented TrueType outline and PostScript outline demonstrate better generalization.
    }
    \label{fig:weight_loss}
\end{figure}

Figure~\ref{fig:weight_loss} presents the training and validation losses for all four outline formats.
Overfitting is observed for original TrueType outline and decomposed TrueType outline, whereas segmented TrueType outline and PostScript outline generalize better without overfitting.

\begin{table}[!t]
    \renewcommand{\arraystretch}{1.3}
    \caption{
        \normalfont
        Performance comparison of different outline formats for font weight classification.
        The results show that the PostScript outline achieves significantly better performance than the TrueType outlines.
    }
    \label{tab:weight_performance}
    \centering
    \begin{tabular}{lcccc}
        \hline
        \textbf{Outline}    & \textbf{Loss} & \textbf{Acc.} & \textbf{Macro F1} & \textbf{W-F1} \\
        \hline
        Original TrueType   & 0.8187        & 69.8\%        & 68.0\%            & 69.4\%        \\
        Decomposed TrueType & 0.7729        & 65.2\%        & 63.6\%            & 64.3\%        \\
        Segmented TrueType  & 0.4274        & 81.1\%        & 79.7\%            & 81.0\%        \\
        PostScript          & 0.3926        & 83.8\%        & 83.9\%            & 83.7\%        \\
        \hline
    \end{tabular}
\end{table}

\section{Conclusion}

In summary, we compared the embedding representations of TrueType and PostScript outlines in Transformer-based vector font classification tasks.
We demonstrated that representations based on PostScript outlines consistently outperform those based on TrueType outlines in font classification accuracy.
The use of PostScript outlines enables a more efficient compression of information in command sequences, enhancing the effect of information aggregation within the Transformer architecture.
This suggests that the choice of font representation is a crucial factor in the performance of deep learning models for vector fonts.

From our results, we found three major advances in the present investigation.
First, the Transformer-based vector font classification model is applicable to complex Kanji character classification and weight classification.
Second, font representations based on PostScript outlines consistently achieve higher classification accuracy than those based on TrueType outlines.
Third, this performance difference is primarily attributed to the segmentation from point sequences to command sequences.

The results show that information aggregation plays a critical role in Transformer-based deep learning for vector graphics.
This role is analogous to tokenization in natural language processing and patch division in computer vision.

Future research should explore further optimization of outline representations.
In particular, we propose adopting the concept of patch division from computer vision and grouping multiple commands into a single token.
This approach could enable better local feature aggregation, leveraging the full potential of the Transformer's information processing capability.
It will be essential to evaluate whether this method improves not only font classification but also font generation and style transfer tasks.

\section*{Acknowledgment}
This work was partly supported by JSPS KAKENHI Grant Number JP23K28154 (GT) and JST CREST Grant Number JPMJCR24R2 (GT).

\section*{Code Availability}

The source code used in this study is available at:

\begin{center}
    \url{https://github.com/fjktkm/truetype-vs-postscript-transformer/}
\end{center}


\end{document}